\documentclass{article}

\PassOptionsToPackage{numbers, compress}{natbib}

\usepackage[preprint]{neurips_2026}

\usepackage[utf8]{inputenc} 
\usepackage[T1]{fontenc}    
\usepackage{hyperref}       
\usepackage{url}            
\usepackage{booktabs}       
\usepackage{amsfonts}       
\usepackage{nicefrac}       
\usepackage{microtype}      
\usepackage{xcolor}         
\usepackage{enumitem}
\usepackage{graphicx}
\usepackage{amsmath}
\usepackage{booktabs}
\usepackage{multirow}
\usepackage{caption}
\usepackage{amssymb}
\usepackage{pifont}

\title{VersaCamVLA: Camera-Configurable \\ VLA Policies for Robotic Manipulation}

\author{
  \parbox{\textwidth}{\centering
    \vspace*{0.3cm}
    Boyao Han$^{1,2*}$ \quad
    Chen Shi$^{1*}$ \quad
    Jingjing Qian$^1$ \quad
    Zhuotao Tian$^{2,3}$ \quad
    Li Jiang$^{1,2\dagger}$ \\[1ex]
    {\normalfont \small
    $^1$The Chinese University of Hong Kong, Shenzhen \\
    $^2$Shenzhen Loop Area Institute \quad
    $^3$Harbin Institute of Technology, Shenzhen} \\[2ex]
    {\normalfont \textbf{Project Page:}
    \href{https://boyaohan.github.io/VersaCamVLA.github.io/}{\small\texttt{https://boyaohan.github.io/VersaCamVLA.github.io/}}}
  }
}

\begin{document}

\maketitle

\begingroup
\renewcommand{\thefootnote}{}
\footnotetext{*: Equal Contribution. \quad $\dagger$: Corresponding author.}
\endgroup

\begin{abstract}
    Vision-Language-Action (VLA) models have emerged as powerful foundations for robotic manipulation, but their reliance on fixed camera configurations during training makes them brittle to changes in camera count or pose during deployment.
    To overcome these limitations, we propose \textbf{VersaCamVLA}, a camera-configurable framework that decouples camera-set representation from action learning. 
    VersaCamVLA learns a unified scene-token interface that maps an arbitrary, variable set of posed RGB views into fixed-size latent scene tokens. This is achieved via multi-signal target-view prediction and Wrist-Augmented Pose Sampling (WAPS), which leverages natural wrist-camera motion for free pose diversity. At deployment, a lightweight spatial encoder injects these compact scene tokens into a pretrained base VLA as a supplementary visual condition, requiring no explicit 3D sensing or novel-view rendering. 
    Experiments on RoboTwin, LIBERO, and a real-robot platform demonstrate that VersaCamVLA consistently outperforms prior VLA methods and direct multi-view baselines, maintaining robust performance across varying camera counts and unseen camera poses.
\end{abstract}

\section{Introduction}

\begin{figure}[t]
  \centering
  \includegraphics[width=1.0\linewidth]{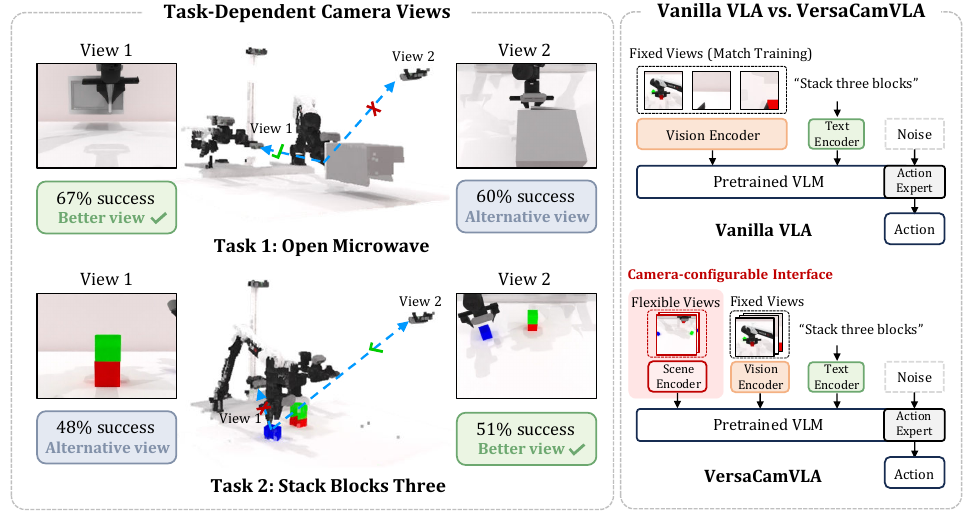}
  \vspace{-2mm}
\caption{
\textbf{Left:} Different manipulation tasks prefer different camera views, making a fixed view setup suboptimal.
\textbf{Right:} VersaCamVLA augments a vanilla VLA with a camera-configurable scene-token interface that enables flexible view setup at deployment.
}
  \label{fig:intro}
  \vspace{-6mm}
\end{figure}

Vision-Language-Action (VLA) models have become a powerful foundation for robotic manipulation, adapting pretrained Vision-Language Models (VLMs) to map visual observations, language instructions, and proprioception to robot actions~\cite{intelligence2025pi_05, black2024pi_0, kim2025openvla-oft, rt12022arxiv, zitkovich2023rt2, octo_2023, nvidia2025gr00tn1openfoundation, pmlr-v270-kim25c}, and their semantic priors and few-shot adaptability make them attractive as general policies across diverse tasks and environments.
However, fine-grained tasks such as precise insertion, and bimanual assembly demand reliable spatial perception, which is tightly coupled to the cameras used during training: changing the number of cameras or shifting a camera's pose alters the visual token distribution and degrades the policy.
This poses a obstacle for deployment, where camera placement is often constrained by the robot, workspace, or task, and a useful view on one platform may be unavailable or suboptimal on another, as illustrated in Fig.~\ref{fig:intro}.


Prior work has addressed camera sensitivity through several complementary directions.
One line of work improves viewpoint robustness by collecting demonstrations from multiple camera poses~\cite{jiang2025you,bian2025lpaf, khazatsky2024droid, xie2026multicameraviewscalingdataefficient} or synthesizing novel views for training-time augmentation~\cite{zhou2023nerfpalmhandcorrective} and training-free test-time adaptation~\cite{heo2026anycamvlazeroshotcameraadaptation}.
However, multi-pose collection is costly on real robots, and synthesis-based methods depend on the fidelity of generated views.
Another line introduces geometric priors through depth, point clouds, or rendered views~\cite{ze2024dp3,liu2026activevla,qian2025geopredict,wilcox2025adapt3radaptive3dscene}, as well as geometric features extracted from RGB images by pretrained 3D foundation models~\cite{abouzeid2026geoawarevlaimplicitgeometryaware, li2025spatialforcingimplicitspatial}.
Finally, multi-camera policies often append visual tokens from additional cameras to the policy input~\cite{zhao2023act,lin2025sem, fan2026peafowl}.
This direct multi-view formulation improves coverage, but it still binds the policy to the camera count and view layout used during action learning, and its token length grows with the number of input views.

We argue that a VLA should instead support a camera-configurable visual pathway that accepts a variable number of posed RGB views, makes no assumption on camera poses, and presents the action policy with a unified fixed-size scene representation.
To this end, we present \textbf{VersaCamVLA}, a camera-configurable VLA framework that decouples camera-set representation learning from action learning.
Rather than redesigning the VLA backbone or requiring explicit 3D reconstruction, VersaCamVLA learns a unified scene-token interface between raw camera observations and the policy.
This interface maps any non-empty set of posed RGB views into fixed-size scene tokens, allowing different camera configurations to be represented through the same policy-facing visual format.

VersaCamVLA is trained in two stages: unified scene-token interface learning and scene-token-conditioned VLA policy learning.
In Stage I, a scene encoder aggregates a source set of posed RGB views into latent scene tokens, and a training-only scene decoder queries these tokens with target-camera rays predict RGB, semantic, and edge signals at target views.
This multi-signal target-view objective encourages the scene tokens to preserve view-consistent spatial information while providing object-level and boundary-aware supervision beyond raw pixel reconstruction.
To expose the encoder to diverse camera configurations without collecting additional multi-pose demonstrations, we introduce \emph{Wrist-Augmented Pose Sampling} (WAPS), which turns the natural motion of wrist cameras into a free source of pose diversity.
At each training frame, WAPS varies the source camera subset while asking the decoder to reconstruct all available views, producing both self-reconstruction and cross-view prediction signals for the same scene tokens.

In Stage II, VersaCamVLA discards the decoder and prediction heads, freezes the scene encoder, and integrates the learned scene tokens into a pretrained VLA policy. The base VLA keeps its native fixed-view RGB input and action-learning objective, while a lightweight spatial encoder compresses the learned scene tokens into a compact supplementary visual condition. Camera flexibility therefore enters through the auxiliary scene-token pathway: any available posed camera subset can be encoded into the same fixed-size representation, while the action expert receives a stable conditioning interface. At deployment, VersaCamVLA supports versatile camera setups and requires only RGB images with camera intrinsics and extrinsics. No novel-view rendering, depth estimation, point-cloud processing, or explicit 3D reconstruction is needed.

We evaluate VersaCamVLA on RoboTwin~\cite{chen2025robotwin}, LIBERO~\cite{liu2023libero}, and a real-robot platform with varying camera counts and poses.
Across these scenarios, VersaCamVLA consistently outperforms prior works and direct multi-view baselines.
These findings show that learning a unified scene-token interface effectively makes pretrained VLA policies more robust to practical 
camera-configuration changes.
\noindent In summary, our main contributions are threefold:
\begin{itemize}[leftmargin=*]
    \item We formulate camera-configurable VLA policy learning, where a policy must operate with a variable number of posed RGB views without assuming fixed camera poses, while maintaining a fixed-size visual interface for action prediction.

    \item We propose VersaCamVLA, a two-stage framework that learns a unified scene-token interface through target-view prediction, 
    then injects the resulting scene tokens into a pretrained VLA as supplementary visual conditioning. Wrist-Augmented Pose Sampling is introduced to exploit wrist-camera motion as free pose diversity for scene-token learning.

    \item We validate VersaCamVLA on simulation and real-robot manipulation benchmarks, showing consistent improvements over prior VLA methods and direct multi-view baselines under changing camera counts and unseen camera poses.
\end{itemize}

\section{Related Work}
\subsection{Vision-Language-Action Models}
Vision-Language-Action models (VLAs)~\cite{intelligence2025pi_05, black2024pi_0, kim2025openvla-oft,lin2025sem, octo_2023, rt12022arxiv, zitkovich2023rt2, li2024cogactfoundationalvisionlanguageactionmodel} adapt pretrained vision-language models (VLMs) to robot control by training on large-scale teleoperation and egocentric data, generalizing across novel scenes, objects, and instructions.
However, recent evaluations show that VLAs are sensitive to camera-configuration shifts, especially viewpoint variations~\cite{fei2025liberoplus,gao2026taxonomy, Wang_2025}, mainly because their 2D RGB inputs lack explicit camera-pose awareness.
Most VLA policies further fix the number of input cameras during post-training, limiting adaptation to camera additions, removals, or pose changes.
Our work targets this underexplored camera-configuration flexibility for pretrained VLAs.

\subsection{Viewpoint Generalization for Vision-Language-Action Policies}

Recent work addresses viewpoint brittleness in visuomotor and Vision-Language-Action policies through data augmentation, test-time view synthesis, and geometry-aware representation learning.
Data augmentation broadens viewpoint coverage through multi-pose demonstrations~\cite{jiang2025you,bian2025lpaf, xie2026multicameraviewscalingdataefficient} or novel-view synthesis~\cite{tian2025vista, chen2024roviaug}, but is limited by collection cost and synthesis fidelity.
Geometry-aware methods inject priors such as depth \cite{qian2025geopredict, zhang20254dvlaspatiotemporalvisionlanguageactionpretraining}, point clouds\cite{huang2026pointworldscaling3dworld, ze2024dp3, peri2024pointcloudmodelsimprove}, or view-aligned 3D features~\cite{ze2024dp3,liu2026activevla,qian2025geopredict,qu2025spatialvla,lu2025gaussianworldmodel, dreamvla25, liu2026activevla}, as well as geometric features extracted from RGB images by pretrained 3D foundation models~\cite{abouzeid2026geoawarevlaimplicitgeometryaware, li2025spatialforcingimplicitspatial}.
In contrast, our method learns compact scene tokens from posed RGB views, improving unseen-pose robustness without explicit 3D reconstruction at deployment or extra multi-pose demonstrations, while requiring calibrated camera intrinsics and extrinsics.

\subsection{Multi-Camera Perception for Vision-Language-Action Policies}
Multi-camera perception is a common strategy in imitation learning and robot manipulation.
Prior policies \cite{zhao2023act, lin2025sem} concatenate image features or tokens from multiple cameras and feed them into a transformer-based action predictor.
Recent VLA-style approaches ~\cite{pmlr-v229-goyal23a, lin2025sem, fan2026peafowl, li2026bfahierarchicalbestfeatureawaretoken, pang2025reviwo, Qian_2025_CVPR, gervet2023act3d3dfeaturefield, guhur2022instructiondrivenhistoryawarepoliciesrobotic} adopt a similar design by appending external-view tokens to the policy input.
This direct multi-view formulation is simple and effective, and has been widely used to improve scene coverage, reduce occlusion, and enhance policy performance on fine-grained manipulation tasks.

\section{Methodology}
\label{sec:method}

\begin{figure}[h]
  \centering
  \includegraphics[width=1.0\linewidth]{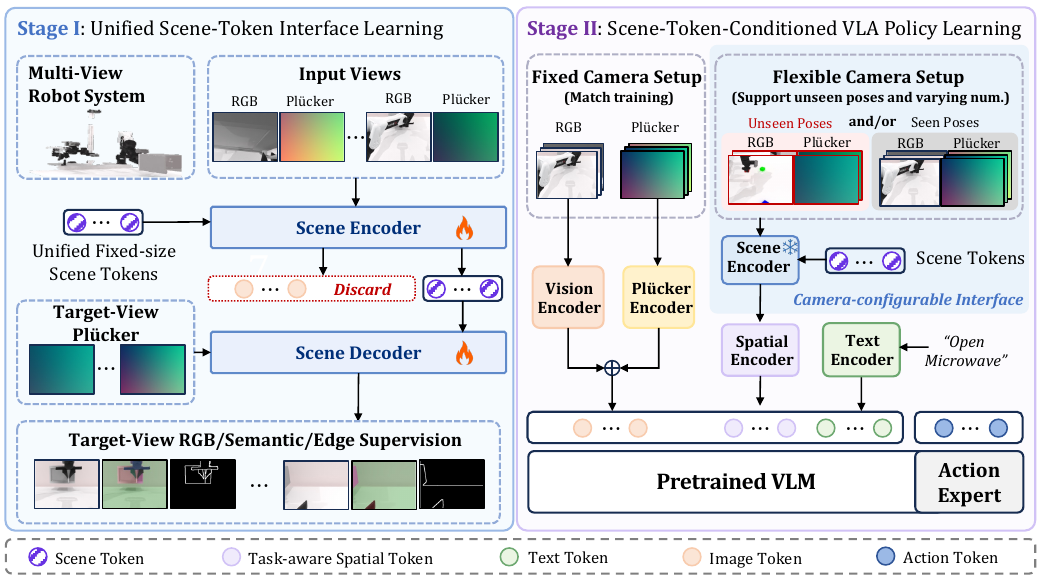}
  \caption{
  \textbf{Overview of VersaCamVLA.}
  \textbf{Stage I:} A scene encoder aggregates posed RGB views into fixed-size scene tokens, with a training-only decoder predicting RGB, semantic, and edge signals at target views.
    \textbf{Stage II:} The decoder is discarded, the scene encoder is frozen, and a jointly-trained spatial encoder compresses the tokens into a supplementary visual condition for a pretrained VLA.
  }
  \vspace{-5mm}
  \label{fig:method}
\end{figure}

We present \textbf{VersaCamVLA}, a camera-configurable VLA framework that decouples camera-set representation learning from action learning, as illustrated in Fig.~\ref{fig:method}.
In Stage I, VersaCamVLA learns a unified scene-token interface between raw observations and the policy. A scene encoder maps any non-empty set of posed RGB views to fixed-size latent scene tokens, supervised by target-view prediction, which encourages the scene tokens to preserve view-consistent spatial information across camera poses.
In Stage II, the decoder used for scene representation learning is removed, the scene encoder is frozen, and a compact version of the learned scene tokens is injected into a pretrained VLA policy as additional visual conditioning, allowing the policy to use observations from different numbers of cameras and unseen camera poses.

%
%
%
Specifically, we first introduce the camera-configurable VLA setting and base policy in Sec.~\ref{sec:problem-setup}.
We then describe how VersaCamVLA learns the unified scene-token interface in Sec.~\ref{sec:scene-encoder-decoder}.
Sec.~\ref{sec:camera-sampling} presents Wrist-Augmented Pose Sampling, which improves camera-configuration coverage during scene representation learning.
Finally, Sec.~\ref{sec:training-vla} describes how the learned scene tokens are integrated into the VLA policy for action learning.

\subsection{Camera-Configurable VLA Setting}
\label{sec:problem-setup}

\noindent\textbf{Standard VLA formulation.}
A VLA policy $\pi$  predicts robot actions at timestep $t$ from a language instruction $\mathcal{L}$, visual observation $\mathcal{V}_t$, and proprioceptive state $\mathcal{P}_t$, written as
\begin{equation}
    A_t=\pi(\mathcal{L}, \mathcal{V}_t, \mathcal{P}_t)
    =[a_t,a_{t+1},\ldots,a_{t+H_a-1}],
\end{equation}
where $H_a$ is the action horizon. Each action $a_t=[\mathbf{q}^{L}_t,g^{L}_t,\mathbf{q}^{R}_t,g^{R}_t]\in\mathbb{R}^{14}$ is represented in joint space: $\mathbf{q}^{L}_t,\mathbf{q}^{R}_t\in\mathbb{R}^{6}$ are the left and right arm joint angles, and $g^{L}_t,g^{R}_t\in\mathbb{R}$ are the gripper states.
In conventional single- or multi-view VLAs, $\mathcal{V}_t$ is tied to the camera setup used during policy training.
For example, a multi-camera VLA typically instantiates $\mathcal{V}_t$ as an image tuple $\{\mathbf{I}_t^i\}_{i=1}^{N}$, where the number of views and the camera poses are fixed components of the policy input.

\vspace{1mm}
\noindent\textbf{Camera-configurable observations.}
The fixed-tuple visual observation in standard VLAs makes the policy interface depend on the demonstrated camera setup.
If a camera is added or moved to a new pose, the resulting visual tokens no longer match the structure seen during action learning.
To remove this dependency, we keep the standard VLA input-output form but replace the camera component of $\mathcal{V}_t$ with a flexible camera set.
Let
$\mathcal{C}=\{(\mathbf{K}^{i},\mathbf{T}^{i})\}_{i=1}^{N}$
denote a camera configuration, where $\mathbf{K}^{i}$ and $\mathbf{T}^{i}$ are the intrinsic and extrinsic matrices of camera $i$.
Under this configuration, the visual observation is $\mathcal{V}_t^{\mathcal{C}}=\{(\mathbf{I}_t^i,\mathbf{K}^{i},\mathbf{T}^{i})\}_{i=1}^{N}$.
Camera-configurable VLA policy learning asks for a single policy that remains valid as $\mathcal{C}$ varies across deployment setups.
The visual observation should therefore allow a variable number of views from non-fixed poses, while the downstream action policy should still receive fixed-size scene tokens.
This motivates VersaCamVLA's scene-token interface, which maps $\mathcal{V}_t^{\mathcal{C}}$ to unified scene tokens $\mathbf{Z}_t$ whose size is independent of $N$.

\vspace{1mm}
\noindent\textbf{Backbone instantiation.}
We instantiate $\pi$ with $\pi_{0.5}$~\cite{intelligence2025pi_05}, a pretrained VLA that uses a VLM~\cite{beyer2024paligemmaversatile3bvlm, bai2023qwenvlversatilevisionlanguagemodel} to encode visual and language context and a robotics-specific action expert to generate continuous action chunks.
VersaCamVLA does not redesign the VLA backbone; instead, it augments the backbone's visual context with scene tokens produced from $\mathcal{V}_t^{\mathcal{C}}$.
The action expert follows $\pi_{0.5}$~\cite{black2024pi_0} and predicts $A_t$ through a flow-matching denoising objective, detailed in Sec.~\ref{sec:training-vla}.

\subsection{Unified Scene-Token Interface Learning}
\label{sec:scene-encoder-decoder}

To implement a unified scene-token interface, VersaCamVLA uses a \textbf{scene encoder} 
that turns the camera-configurable observation 
into fixed-size scene tokens $\mathbf{Z}$.
Inspired by LVSM~\cite{jin2024lvsm}, we learn this encoder with target-view prediction: a \textbf{scene decoder}, used only during representation learning, queries $\mathbf{Z}$ with target-camera rays and reconstructs RGB, semantic, and edge signals.
This objective encourages $\mathbf{Z}$ to support prediction at target views beyond the input views, while presenting the downstream VLA with a consistent visual representation regardless of the input camera set.



\noindent\textbf{Scene encoder.}
Given a source camera set $\mathcal{S}$ and the corresponding input observation $\mathcal{V}^{\mathcal{S}}=\{(\mathbf{I}^i,\mathbf{K}^{i},\mathbf{T}^{i})\}_{i=1}^{N}$ with $N$ source views, we first compute a Pl\"ucker ray embedding map $\mathbf{R}^{i}$ for each posed view.
We then tokenize each RGB image and ray map via ViT-style patchification with patch size $p$, yielding $P=H_IW_I/p^{2}$ patches per view.
For the $j$-th patch of view $i$, we concatenate the RGB patch $\mathbf{I}^{i,j}\in\mathbb{R}^{p\times p\times 3}$ with its ray embedding patch $\mathbf{R}^{i,j}\in\mathbb{R}^{p\times p\times 6}$, and project the result into a $d$-dimensional source token:
\begin{equation}
    \mathbf{x}^{i,j} = \phi_{\mathrm{src}}\!\left(\left[\mathbf{I}^{i,j},\,\mathbf{R}^{i,j}\right]\right) \in \mathbb{R}^{d},
\end{equation}
where \(\phi_{\mathrm{src}}\) is a linear patch projection.
The tokens from all source views are stacked into a source sequence $\mathbf{X}\in\mathbb{R}^{(N\times P)\times d}$, whose length varies with the number of input cameras.

We use $L$ learnable scene tokens $\mathbf{E}=\{\mathbf{e}^{k}\}_{k=1}^{L}$ with $\mathbf{e}^{k}\in\mathbb{R}^{d}$ to aggregate the variable-length source-token sequence into a fixed-size representation.
The source tokens and scene tokens are concatenated and processed jointly by a Transformer encoder, allowing the scene tokens to attend to spatial information from all input views through self-attention.
We discard the updated source tokens and retain the updated scene-token outputs $\mathbf{Z}$ as the compact scene representation:
\begin{equation}
    \mathbf{Z} = \mathrm{Enc}_{\theta}\!\left(\left[\mathbf{X},\,\mathbf{E}\right]\right) \in \mathbb{R}^{L\times d}.
\end{equation}
Since \(L\) is independent of the number and poses of input cameras, \(\mathbf{Z}\) provides a fixed-size scene representation for downstream policy learning under flexible camera configurations.

\vspace{1mm}
\noindent\textbf{Scene decoder.}
To encourage $\mathbf{Z}$ to encode view-consistent scene information that remains useful beyond the input camera views,
we train a scene decoder to predict signals at target view poses conditioned on $\mathbf{Z}$.
Given a target camera $c=(\mathbf{K},\mathbf{T})$, we denote its tokenized Pl\"ucker ray embeddings as $\{\mathbf{R}^{c,j}\}_{j=1}^{P}$ and project each target ray patch through a linear projection $\phi_{\mathrm{tgt}}$:
\begin{equation}
    \mathbf{q}^{c,j} = \phi_{\mathrm{tgt}}\!\left(\mathbf{R}^{c,j}\right) \in \mathbb{R}^{d}.
\end{equation}
The target ray tokens $\mathbf{Q}^{c}\in\mathbb{R}^{P\times d}$ are concatenated with the scene tokens and passed through a Transformer with self-attention. We keep only the updated target tokens $\mathbf{Y}^{c}$ as output:
\begin{equation}
    \mathbf{Y}^{c} = \mathrm{Dec}_{\psi}\!\left(\left[\mathbf{Q}^{c},\,\mathbf{Z}\right]\right) \in \mathbb{R}^{P\times d}.
\end{equation}
$\mathbf{Y}^{c}$ are then fed to lightweight prediction heads for RGB, semantic-map, and edge-map reconstruction.
Successful prediction requires $\mathbf{Z}$ to summarize the source observations in a form that supports prediction beyond the encoder input views.


\textbf{Multi-signal target-view supervision.}
Given a target camera set $\mathcal{T}$, the scene decoder produces target-view output tokens $\mathbf{Y}^{c}$ for each $c\in\mathcal{T}$.
A lightweight prediction head $h_{\mathrm{rgb}}$ then maps $\mathbf{Y}^{c}$ to a predicted RGB image:
\begin{equation}
    \hat{\mathbf{I}}^{c} = h_{\mathrm{rgb}}(\mathbf{Y}^{c}),
\end{equation}
and is supervised against the ground-truth RGB image $\mathbf{I}^{c}$ by an MSE term plus a deep perceptual term $\mathcal{L}_{\mathrm{perc}}$ computed in a pretrained feature space:
\begin{equation}
    \mathcal{L}_{\mathrm{rgb}} =
    \sum_{c\in\mathcal{T}}
    \left[
    \mathrm{MSE}(\hat{\mathbf{I}}^{c}, \mathbf{I}^{c}) +
    \lambda_{\mathrm{perc}}\,\mathcal{L}_{\mathrm{perc}}(\hat{\mathbf{I}}^{c}, \mathbf{I}^{c})
    \right].
\end{equation}

RGB reconstruction alone can be dominated by texture, lighting, and background statistics.
To better align the scene tokens with manipulation-relevant structure, we add two complementary supervision signals: semantic maps and edge maps.
Two additional heads on the same target-view decoder output produce the semantic and edge predictions:
\begin{equation}
    \hat{\mathbf{S}}^{c} = h_{\mathrm{sem}}(\mathbf{Y}^{c}),
    \qquad
    \hat{\mathbf{B}}^{c} = h_{\mathrm{edge}}(\mathbf{Y}^{c}),
\end{equation}
each supervised against its target map by an MSE term:
\begin{equation}
    \mathcal{L}_{\mathrm{sem}} = \sum_{c\in\mathcal{T}} \mathrm{MSE}(\hat{\mathbf{S}}^{c}, \mathbf{S}^{c}),
    \qquad
    \mathcal{L}_{\mathrm{edge}} = \sum_{c\in\mathcal{T}} \mathrm{MSE}(\hat{\mathbf{B}}^{c}, \mathbf{B}^{c}).
\end{equation}
Semantic targets $\mathbf{S}^{c}$ come from the simulator's segmentation sensor in simulation and from a SAM3 segmenter \cite{carion2025sam3} fine-tuned on our real-robot data for real frames; edge targets $\mathbf{B}^{c}$ are extracted with a Canny detector.
The scene-token learning objective is the weighted sum:
\begin{equation}
    \mathcal{L}_{\mathrm{scene}} = \mathcal{L}_{\mathrm{rgb}} + \lambda_{\mathrm{sem}}\,\mathcal{L}_{\mathrm{sem}} + \lambda_{\mathrm{edge}}\,\mathcal{L}_{\mathrm{edge}},
\end{equation}
where $\lambda_{\mathrm{sem}}$ and $\lambda_{\mathrm{edge}}$ balance the high-level signals against photometric reconstruction.

\subsection{Wrist-Augmented Pose Sampling}
\label{sec:camera-sampling}

The scene encoder in Sec.~\ref{sec:scene-encoder-decoder} is designed to accept a variable number of views from non-fixed poses and produce fixed-size unified scene tokens for downstream VLA.
Learning this camera-configurable interface, however, requires the training data to cover sufficiently diverse camera placements.
In tabletop manipulation data, static cameras provide stable but sparsely distributed viewpoints, while wrist cameras sweep a much richer pose distribution as the robot moves.
Wrist-Augmented Pose Sampling (WAPS) turns this natural motion into pose diversity for scene-token learning.

Specifically, for each training frame, let $\mathcal{C}_{\mathrm{stat}}$ and $\mathcal{C}_{\mathrm{wrist}}$ denote the available static-camera and wrist-camera views.
The source view set $\mathcal{S}=\mathcal{S}_{\mathrm{stat}}\cup\mathcal{S}_{\mathrm{wrist}}$ fed to the scene encoder is drawn from the two camera types independently.
From the static cameras, we randomly sample $n$ views as $\mathcal{S}_{\mathrm{stat}}$, where $n$ is a random integer between $1$ and $|\mathcal{C}_{\mathrm{stat}}|$.
For wrist cameras, each view is independently included in $\mathcal{S}_{\mathrm{wrist}}$ with probability $0.5$.
%
We use the full camera pool as the target set $\mathcal{T}=\mathcal{C}_{\mathrm{stat}}\cup\mathcal{C}_{\mathrm{wrist}}$, so the decoder is always trained to render every available view, regardless of the encoder input.

\begin{figure}[t]
  \centering
  \includegraphics[width=1.0\linewidth]{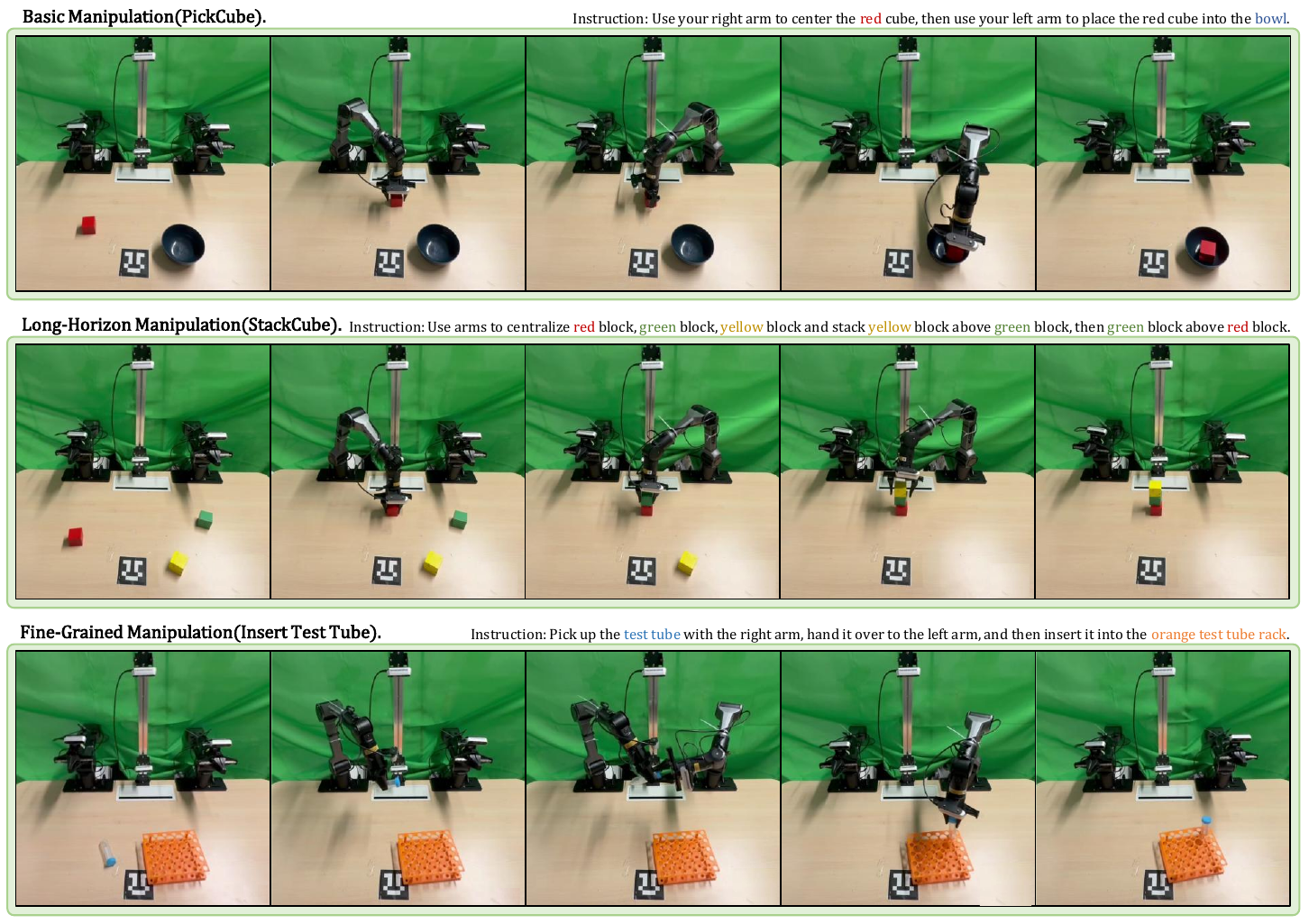}
  \caption{
    \textbf{Real-world Manipulation Processes.} 
        Representative rollouts of VersaCamVLA on three real-world tasks: PickCube, StackCube, and Insert Test Tube, showing key execution states from the initial scene to task completion.
  }
  \vspace{-5mm}
  \label{fig:test-tube}
\end{figure}

This design provides two complementary supervision signals on the same scene tokens.
Self-reconstruction on source views preserves visual detail through the latent bottleneck, while cross-view prediction on held-out views forces $\mathbf{Z}$ to support view poses that were not in the encoder input.
By varying the number and poses of source cameras across training frames, the encoder learns to extract scene representation from arbitrary camera subsets instead of relying on a fixed camera setup. 
Wrist-camera motion further provides diverse cross-view targets at no extra data-collection cost.



\begin{table}[t]
  \caption{
        \textbf{Simulation results on RoboTwin 2.0.}
        We report success rates (\%) on RoboTwin 2.0.
       All results are reproduced by us. \textbf{Bold} indicates the best performance in each column.
    }
  \vspace{1mm}
  \label{tab:robotwin_clean_dr}
  \centering
  \small
  \setlength{\tabcolsep}{3pt}
  \renewcommand{\arraystretch}{0.99}
  \resizebox{\linewidth}{!}{
  \begin{tabular}{@{}lc*{6}{cc}@{}}
    \toprule
    \multirow{2}{*}{Task}
    & \multicolumn{2}{c}{ACT\cite{zhao2023act}}
    & \multicolumn{2}{c}{DP\cite{chi2024dp}}
    & \multicolumn{2}{c}{DP3\cite{ze2024dp3}}
    & \multicolumn{2}{c}{OpenVLA-OFT\cite{kim2025openvla-oft}}
    & \multicolumn{2}{c}{$\pi_{0.5}$ \cite{intelligence2025pi_05}}
    & \multicolumn{2}{c}{VersaCamVLA} \\
    \cmidrule(lr){2-3}
    \cmidrule(lr){4-5}
    \cmidrule(lr){6-7}
    \cmidrule(lr){8-9}
    \cmidrule(lr){10-11}
    \cmidrule(lr){12-13}
    & Clean & DR
    & Clean & DR
    & Clean & DR
    & Clean & DR
    & Clean & DR
    & Clean & DR \\
    \midrule
    Blocks Ranking RGB      & 0  & 0 & 0  & 0 & 2  & 0 & 0  & 0 & 38 & 11 & \textbf{67} & \textbf{12} \\
    Blocks Ranking Size     & 3  & 0 & 1  & 0 & 2  & 1 & 7  & 0 & 12 & 4  & \textbf{35} & \textbf{6} \\
    Hanging Mug             & 6  & 0 & 19 & 0 & \textbf{35} & 1 & 12 & 0 & 4  & \textbf{4}  & 10 & \textbf{4} \\
    Move Stapler Pad        & 0  & 0 & 0  & 0 & 7  & 0 & 0  & 0 & 5  & \textbf{7} & \textbf{8} & 5 \\
    Open Laptop             & 72 & 1 & 53 & 0 & 77 & 1 & 82 & 0 & 90 & \textbf{70} & \textbf{96} & 62 \\
    Open Microwave          & 72 & 0 & 78 & 0 & \textbf{91} & 25 & 23 & \textbf{56} & 43 & 26 & 51 & 46 \\
    Pick Diverse Bottles    & 8  & 0 & 28 & 0 & \textbf{55} & 1 & 9  & 0 & 36 & \textbf{21} & 47 & 12 \\
    Place A2B Left          & 1  & 0 & 3  & 0 & 30 & 1 & 5  & 0 & 42 & \textbf{24} & \textbf{54} & 17 \\
    Place A2B Right         & 2  & 0 & 6  & 0 & 43 & 0 & 11 & 0 & 24 & 19 & \textbf{45} & \textbf{27} \\
    Place Burger Fries      & 67 & 0 & 80 & 0 & 75 & 2 & 30 & 0 & 48 & 31 & \textbf{81} & \textbf{49} \\
    Place Dual Shoes        & 0  & 0 & 4  & 0 & 9  & 0 & 2  & 0 & 25 & 24 & \textbf{51} & \textbf{25} \\
    Place Empty Cup         & 38 & 0 & 27 & 0 & 75 & 1 & 20 & 0 & 77 & 1  & \textbf{81} & \textbf{27} \\
    Place Object Basket     & 1  & 0 & 21 & 0 & 49 & 3 & 5  & 0 & 53 & 26 & \textbf{76} & \textbf{35} \\
    Scan Object             & 3  & 0 & 8  & 0 & 29 & 1 & 8  & 1 & 6  & 4  & \textbf{35} & \textbf{15} \\
    Stack Blocks Three      & 5  & 0 & 0  & 0 & 3  & 0 & 0  & 0 & 18 & \textbf{50} & \textbf{43} & 3 \\
    Stack Bowls Three       & 37 & 0 & 51 & 0 & 57 & 4 & 29 & 0 & 53 & \textbf{24} & \textbf{61} & 23 \\
    \midrule
    Average (\%)       & 19.69 & 0.06 & 23.69 & 0.00 & 39.94 & 2.56 & 15.19 & 3.56 & 35.88 & 21.63 & \textbf{52.56} & \textbf{23.00} \\
    \bottomrule
  \end{tabular}
  }
  \vspace{-5mm}
\end{table}

\begin{table}[t]
  \caption{
        \textbf{Simulation results on LIBERO.}
        We report task success rates (\%) on LIBERO.
        Results marked with $^\ast$ are reproduced by us.
        \textbf{Bold} indicates the best performance in each column, and \underline{underlined} indicates the second-best performance.
    }
  \vspace{1mm}
  \label{tab:libero_results}
  \centering
  \small
  \setlength{\tabcolsep}{5pt}
  \renewcommand{\arraystretch}{0.99}
  \resizebox{0.9\linewidth}{!}{
  \begin{tabular}{@{}lccccc@{}}
    \toprule
    Method 
    & LIBERO-Spatial 
    & LIBERO-Goal 
    & LIBERO-Object 
    & LIBERO-Long
    & Average \\
    \midrule
    Diffusion Policy\cite{chi2024dp} & 78.3 
    & 92.5 
    & 68.3 
    & 50.5 
    & 72.4 \\
    OpenVLA-OFT \cite{kim2025openvla-oft}
    & 95.2 
    & 94.2 
    & 95.2 
    & \textbf{93.2} 
    & 94.5 \\
    $\pi_{0.5}^{\ast}$ \cite{intelligence2025pi_05}
    & \underline{98.0} 
    & \underline{94.6} 
    & \textbf{99.0} 
    & 91.6 
    & \underline{95.8} \\
    VersaCamVLA 
    & \textbf{98.2} 
    & \textbf{95.0} 
    & \underline{97.8} 
    & \underline{92.4} 
    & \textbf{95.9} \\
    \bottomrule
  \end{tabular}
  }
  \vspace{-5mm}
\end{table}

\subsection{Scene-Token-Conditioned VLA Policy Learning}
\label{sec:training-vla}

After learning the scene-token interface, we discard the scene decoder and prediction heads, freeze the scene encoder, and integrate the resulting scene tokens into the base VLA for action learning.


\vspace{1mm}
\noindent\textbf{Action policy learning.}
At time step $t$, the frozen scene encoder maps the versatile camera configuration $\mathcal{C}$ and the corresponding posed views $\mathcal{V}^{\mathcal{C}}_{t}$ into scene tokens $\mathbf{Z}_{t}$ following Sec.~\ref{sec:scene-encoder-decoder}.
Although $\mathbf{Z}_{t}$ has fixed length, directly feeding all scene tokens into the VLA can still add unnecessary computation and may expose the policy to background details that are irrelevant to action prediction.
A convolutional spatial encoder $g_{\eta}$ compresses scene tokens into a compact, task-oriented representation:
\begin{equation}
    \bar{\mathbf{Z}}_{t} = g_{\eta}(\mathbf{Z}_{t}).
\end{equation}
The compact $\bar{\mathbf{Z}}_{t}$ is fed into VLA as a supplementary scene representation.
The policy therefore conditions on the language instruction $\mathcal{L}$, the RGB observations $\mathcal{V}_{t}$, the supplementary scene representation $\bar{\mathbf{Z}}_{t}$, and the proprioceptive state $\mathcal{P}_{t}$, and predicts the action chunk $A_{t}$.

The scene encoder stays frozen throughout this stage; $g_{\eta}$ is randomly initialized and, jointly with base VLA's trainable parameters $\omega$, optimized by the standard flow-matching action regression loss:
\begin{equation}
    \mathcal{L}_{\mathrm{action}}
    =
    \mathbb{E}_{\tau,\epsilon}\!\left[
        \left\|
            v_{\eta,\omega}\!\left(A_{t}^{\tau};\, \mathcal{L}, \mathcal{V}_{t}, \bar{\mathbf{Z}}_{t}, \mathcal{P}_{t}, \tau\right)
            - (A_{t}-\epsilon)
        \right\|_{2}^{2}
    \right],
\end{equation}
where $\epsilon\sim\mathcal{N}(0,\mathbf{I})$, $\tau\sim\mathcal{U}[0,1]$, $A_{t}^{\tau}=\tau A_{t}+(1-\tau)\epsilon$ is the linear interpolation between noise and the demonstration action chunk, and $v_{\eta,\omega}$ is the velocity field predicted by the base VLA $\pi_{0.5}$'s action expert.
Because $g_{\eta}$ is supervised only through $\mathcal{L}_{\mathrm{action}}$, it acts as a task-driven information bottleneck that retains scene-token information useful for action prediction.


\begin{table}[!t]
  \centering
  \small
  \renewcommand{\arraystretch}{0.99}
  \caption{
    \textbf{Robustness to camera-pose variations.}
    Average success rate (\%) on RoboTwin~2.0 \textit{Clean} and real-robot tasks.
    \textbf{Bold} indicates the best performance in each column.
  }
  \vspace{1mm}
  \label{tab:camera_pose_ablation}
  \setlength{\tabcolsep}{5pt}

  \begin{tabular*}{0.9\linewidth}
    {@{\extracolsep{\fill}}lcccc@{}}
    \toprule
    & \multicolumn{2}{c}{RoboTwin~2.0 }
    & \multicolumn{2}{c}{Real Robot} \\
    \cmidrule(lr){2-3}
    \cmidrule(lr){4-5}
    Method
    & Seen Pose
    & Unseen Pose
    & Seen Pose
    & Unseen Pose \\
    \midrule

    $\pi_{0.5}$
    & 48.88
    & 32.88
    & 41.67
    & 13.33 \\

    VersaCamVLA
    & \textbf{53.44}
    & \textbf{52.63}
    & \textbf{53.33}
    & \textbf{58.33} \\

    \bottomrule
  \end{tabular*}

  \vspace{-4mm}
\end{table}

\vspace{-2mm}

\begin{table}[!t]
  \centering
  \small
  \renewcommand{\arraystretch}{0.99}
  \caption{
    \textbf{Effect of input camera count.}
    Average success rate (\%) on RoboTwin~2.0 \textit{Clean} and real-robot tasks.
    V denotes views.
    \textbf{Bold} indicates the best performance in each column.
  }
  \vspace{1mm}
  \label{tab:camera_count_ablation}
  \setlength{\tabcolsep}{5pt}

  \begin{tabular*}{0.9\linewidth}
    {@{\extracolsep{\fill}}lccccccc@{}}
    \toprule
    & \multicolumn{4}{c}{RoboTwin~2.0}
    & \multicolumn{3}{c}{Real Robot} \\
    \cmidrule(lr){2-5}
    \cmidrule(lr){6-8}
    Method
    & 3V
    & 4V
    & 5V
    & 6V
    & 3V
    & 4V
    & 5V \\
    \midrule

    $\pi_{0.5}$
    & 35.88
    & 48.88
    & \ding{56}
    & \ding{56}
    & \textbf{31.67}
    & 41.67
    & \ding{56} \\

    VersaCamVLA
    & \textbf{52.56}
    & \textbf{53.44}
    & \textbf{53.13}
    & \textbf{55.63}
    & \ding{56}
    & \textbf{53.33}
    & \textbf{65.00} \\

    \bottomrule
  \end{tabular*}

  \vspace{-4mm}
\end{table}

\noindent\textbf{Inference.}
At deployment, the scene decoder and the three prediction heads are discarded; only the scene encoder, spatial encoder, and the base VLA backbone (\(\pi_{0.5}\)) run on the inference path.
Any subset of posed camera observations available on the robot system is encoded into the same fixed-length \(\bar{\mathbf{Z}}_{t}\), so the policy operates under flexible camera configurations through a unified scene-token interface, without novel-view rendering, depth, point clouds, or explicit 3D reconstruction.

\section{Experiments}
\label{experiments} 

\subsection{Experimental Setup}
\textbf{Simulation Setup.}
We evaluate on two standard robotic manipulation benchmarks, \textbf{RoboTwin 2.0}~\cite{chen2025robotwin} for bimanual manipulation under canonical (\textit{Clean}) and domain-randomized (\textit{DR}) settings, and \textbf{LIBERO}~\cite{liu2023libero} for single-arm manipulation across four suites.
Embodiments, task selection, demonstration counts, and training protocols are described in Appendix~\ref{app:benchmark_details}.

\textbf{Camera-Configuration Setup.}
Beyond the official benchmark settings, we design additional evaluations to test robustness to changing camera configurations.
\begin{itemize}[leftmargin=*, nosep]
    \item \textbf{Camera-Pose Perturbation.}
    On both RoboTwin 2.0 and LIBERO, policies are trained under the nominal camera setup and evaluated with camera poses perturbed in azimuth, distance, and pitch over a workspace-centered spherical region, probing generalization to unseen viewpoints.

    \item \textbf{Camera-Count Variation.}
    On RoboTwin 2.0 \textit{Clean}, we train two $\pi_{0.5}$ baselines separately at 3 and 4 views and a single VersaCamVLA policy, then evaluate VersaCamVLA under 3, 4, 5, and 6 input views and each $\pi_{0.5}$ baseline at its matched view count, isolating the effect of camera count.
\end{itemize}

\textbf{Real-World Setup.}
To evaluate real-robot camera flexibility, we test VersaCamVLA on three Cobot Magic bimanual tasks, covering sequential coordination, long-horizon manipulation, and fine handover-and-insertion, with 20 trajectories per task. Figure~\ref{fig:test-tube} provides visualizations of all tasks.

\begin{itemize}[leftmargin=*, nosep]
    \item \textbf{Basic Manipulation (PickCube).}
    The right arm moves a red cube to the workspace center, and the left arm then picks it up and places it into a bowl, exercising sequential bimanual coordination in a shared workspace.
    
    \item \textbf{Long-Horizon Manipulation (StackCube).}
    The robot stacks three cubes into a red--green--yellow tower from bottom to top with cube positions randomized at evaluation; small placement errors collapse the tower, stressing precise spatial control over an extended horizon.
    
    \item \textbf{Fine-Grained Manipulation(Insert Test Tube):}
    The right arm grasps a test tube, transfers it to the left arm, and inserts it into a tube rack, requiring accurate handover orientation and fine local alignment despite gripper-induced occlusions during the handover.
\end{itemize}

\vspace{-0.5mm}

\textbf{Baselines.} We compare VersaCamVLA with visuomotor and bimanual VLA baselines. Visuomotor policies include ACT \cite{zhao2023act}, DP \cite{chi2024dp} and DP3 \cite{ze2024dp3} trained separately per task. VLA baselines include OpenVLA-OFT~\cite{kim2025openvla-oft}, the original $\pi_{0.5}$~\cite{intelligence2025pi_05}. For $\pi_{0.5}$, each additional RGB view is encoded with SigLIP \cite{zhai2023sigmoidlosslanguageimage} and appended to the policy input as extra visual tokens in a fixed view order.

\textbf{Implementation Details.}
We use a predictive horizon of $H=50$ for all tasks.
Details of the camera-pose perturbation sampling protocol are provided in Appendix~\ref{app:impl_perturb}.
We train the policy for 30,000 iterations using AdamW~\cite{loshchilov2017adamw} with a learning rate of $5\times10^{-5}$ on 8 NVIDIA H100 GPUs and a total batch size of 512.
The primary evaluation metric is task success rate (\%).More implementation details are provided in Appendix~\ref{app:impl_details}.

\begin{table}[!t]
\centering
\small
\renewcommand{\arraystretch}{0.99}


\begin{minipage}[t]{0.48\linewidth}
  \vspace{0pt}
  \centering
  \caption{
    \textbf{Token ablation on RoboTwin~2.0 \textit{Clean}.}
    Average success rate (\%). \textbf{Bold} indicates the best performance.
  }
  \label{tab:token_ablation_robotwin}
  \vspace{1mm}
  \setlength{\tabcolsep}{4pt}
  \begin{tabular*}{\linewidth}{@{\extracolsep{\fill}}lccc@{}}
    \toprule
    Token Number & 48 & 192 & 432 \\
    \midrule
    Average SR (\%)
    & 50.06
    & \textbf{55.63}
    & 47.50 \\
    \bottomrule
  \end{tabular*}
\end{minipage}
\hfill
\begin{minipage}[t]{0.48\linewidth}
  \vspace{0pt}
  \centering
  \caption{
    \textbf{Spatial Encoder ablation on RoboTwin~2.0 \textit{Clean}.}
    Average success rate (\%). \textbf{Bold} indicates the best performance.
  }
  \label{tab:encoder_ablation_robotwin}
  \vspace{1mm}
  \setlength{\tabcolsep}{4pt}
  \begin{tabular*}{\linewidth}{@{\extracolsep{\fill}}lcc@{}}
    \toprule
    Spatial Encoder & Pooling & Conv \\
    \midrule
    Average SR (\%)
    & 50.13
    & \textbf{55.63} \\
    \bottomrule
  \end{tabular*}
\end{minipage}

\vspace{0mm}


\begin{minipage}[t]{0.48\linewidth}
  \vspace{0pt}
  \centering
  \caption{
    \textbf{WAPS ablation.}
    Average success rate (\%) under the unseen camera pose setting.
    \textbf{Bold} indicates the best result.
  }
  \label{tab:ablation_waps}
  \vspace{1mm}
  \setlength{\tabcolsep}{4pt}
  \begin{tabular*}{\linewidth}{@{\extracolsep{\fill}}lcc@{}}
    \toprule
    Method & w/o WAPS & WAPS \\
    \midrule
    Average SR (\%)
    & 48.79
    & \textbf{52.63} \\
    \bottomrule
  \end{tabular*}
\end{minipage}
\hfill
\begin{minipage}[t]{0.48\linewidth}
  \vspace{0pt}
  \centering
  \caption{
    \textbf{LVSM pretraining ablation.}
    Average success rate (\%) on RoboTwin~2.0 \textit{Clean}.
    \textbf{Bold} indicates the best result.
  }
  \label{tab:ablation_lvsm_pretraining}
  \vspace{1mm}
  \setlength{\tabcolsep}{4pt}
  \begin{tabular*}{\linewidth}{@{\extracolsep{\fill}}lcc@{}}
    \toprule
    Initialization & From scratch & Pretrained LVSM \\
    \midrule
    Average SR (\%)
    & 51.25
    & \textbf{55.63} \\
    \bottomrule
  \end{tabular*}
\end{minipage}

\vspace{-5mm}
\end{table}

\vspace{-1mm}

\subsection{Main Results}

\textbf{Simulation Results.}

\begin{itemize}[leftmargin=*]
    \item \textbf{RoboTwin 2.0.}
    Table~\ref{tab:robotwin_clean_dr} reports per-task success rates on the 16 fine-grained RoboTwin~2.0 tasks under the official \textit{Clean} and DR settings, with 100 trials per task.
    Both $\pi_{0.5}$ and VersaCamVLA use 3-view inputs.
    VersaCamVLA achieves the best average in both settings, reaching 52.56\% on \textit{Clean} and 23.00\% on DR.
    VersaCamVLA achieves the best DR average, showing that scene tokens provide robust policy representations under appearance randomization.

    \item \textbf{LIBERO.}
    Table~\ref{tab:libero_results} reports results on the four LIBERO suites, testing whether the scene-token pathway preserves the general manipulation capability of the pretrained backbone on standard single-arm tasks.
    VersaCamVLA reaches an average success rate of 95.9\%, substantially above Diffusion Policy\cite{chi2024dp} (72.4\%) and OpenVLA-OFT\cite{kim2025openvla-oft} (94.5\%) and on par with the strong $\pi_{0.5}$ \cite{intelligence2025pi_05} baseline (95.8\%), with slight gains on LIBERO-Spatial, LIBERO-Goal, and LIBERO-Long.
\end{itemize}

\vspace{-1mm}

\textbf{Camera-Configuration Results.}

\vspace{-1mm}

\begin{itemize}[leftmargin=*]
    \item \textbf{Camera-Pose Perturbation.}
    Table~\ref{tab:camera_pose_ablation} summarizes camera-pose robustness on both RoboTwin~2.0 and real-robot tasks. The LIBERO counterpart deferred to the Appendix \ref{app:per_task_libero_pose}.
    Both $\pi_{0.5}$ and VersaCamVLA are evaluated under matched four-view inputs, and under perturbation $\pi_{0.5}$ drops 16.0 pp while VersaCamVLA loses only 0.8 pp.
    These results show that the scene-token interface is more pose-robust than direct multi-view token concatenation.

    \item \textbf{Camera-Count Variation.}
    Table~\ref{tab:camera_count_ablation} summarizes the effect of camera-count on both RoboTwin~2.0 and real-robot tasks. On RoboTwin 2.0 \textit{Clean}, we train two $\pi_{0.5}$ baselines separately at 3 and 4 views and a single VersaCamVLA policy, then evaluate VersaCamVLA under 3, 4, 5, and 6 input views and each $\pi_{0.5}$ baseline at its matched view count, isolating the effect of camera count.
\end{itemize}

\vspace{-1mm}

\textbf{Real-world Results.}
Tables~\ref{tab:camera_pose_ablation} and~\ref{tab:camera_count_ablation} report real-robot success rates on three bimanual manipulation tasks; the same VersaCamVLA model is evaluated at the trained pose, at an unseen pose, and with one extra test-time camera, while $\pi_{0.5}$ is trained separately at 3 and 4 views and tested at the matched count.
Under matched 4-view inputs, VersaCamVLA outperforms $\pi_{0.5}$ by 11.7 and 45.0 percentage points on seen and unseen camera poses, respectively.

\vspace{-2mm}

\subsection{Ablation Study}
\label{sec:ablation}
We ablate five design choices in VersaCamVLA, covering how the scene-token interface is learned and how it conditions the action policy.

\textbf{Number of compressed scene tokens.}
Table~\ref{tab:token_ablation_robotwin} varies the number of compressed scene tokens passed to the policy.
Using $192$ tokens achieves the highest average success of $55.63\%$, compared with $50.06\%$ for $48$ tokens and $47.50\%$ for $432$ tokens.

\textbf{Spatial encoder architecture.}
The spatial encoder compresses scene tokens into a supplementary visual condition for the action policy.
As shown in Table~\ref{tab:encoder_ablation_robotwin}, convolutional compression achieves $55.63\%$ average success, compared with $50.13\%$ for pooling.

\textbf{Wrist-Augmented Pose Sampling.}
Learning a camera-configurable interface requires exposure to diverse camera poses.
Table~\ref{tab:ablation_waps} shows that WAPS improves average success under unseen camera poses from $48.79\%$ to $52.63\%$, a gain of $3.84$ percentage points.

\textbf{Scene encoder--decoder initialization.}
Table~\ref{tab:ablation_lvsm_pretraining} examines the contribution of LVSM initialization.
Initializing the scene encoder--decoder from a pretrained LVSM checkpoint achieves $55.63\%$ average success, compared with $51.25\%$ when trained from scratch.

\textbf{Multi-signal supervision.}
RGB reconstruction provides a basic learning signal, while semantic and edge targets introduce object-level and boundary information.
Table~\ref{tab:ablation_signal_summary} shows that adding edge or semantic supervision individually improves average success from $20.94\%$ to $24.75\%$ and $25.69\%$, respectively.
Combining both signals yields the best result of $29.50\%$, an improvement of $8.56$ percentage points over RGB-only supervision.

\begin{table}[t]
  \centering
  \normalsize
  \renewcommand{\arraystretch}{1.1}
  \begin{minipage}[t]{0.99\linewidth}
    \vspace{0pt}
    \centering
    \caption{
      \textbf{Ablation of supervision signals for scene-token learning.}
      Average success rate (\%) on RoboTwin~2.0 DR. \textbf{Bold} indicates the best result.}
    \label{tab:ablation_signal_summary}
    \vspace{1mm}
    \setlength{\tabcolsep}{2pt}
    \begin{tabular*}{\linewidth}{@{\extracolsep{\fill}}lcccc@{}}
      \toprule
      RGB      & \textcolor{red}{\ding{51}} & \textcolor{red}{\ding{51}} & \textcolor{red}{\ding{51}} & \textcolor{red}{\ding{51}} \\
      Edge     & \textcolor{blue}{\ding{55}} & \textcolor{red}{\ding{51}} & \textcolor{blue}{\ding{55}} & \textcolor{red}{\ding{51}} \\
      Semantic & \textcolor{blue}{\ding{55}} & \textcolor{blue}{\ding{55}} & \textcolor{red}{\ding{51}} & \textcolor{red}{\ding{51}} \\
      \midrule
      Average SR & 20.94 & 24.75 & 25.69 & \textbf{29.50} \\
      \bottomrule
    \end{tabular*}
  \end{minipage}
  \vspace{-3mm}
\end{table}

\section{Conclusion}
\label{conclusion}
We present \textbf{VersaCamVLA}, a camera-configurable VLA framework that augments pretrained policies with a unified scene-token interface decoupling camera-set representation from action learning.
VersaCamVLA introduces a scene encoder learned by multi-signal target-view prediction and \emph{Wrist-Augmented Pose Sampling} that exploits wrist-camera motion as free pose diversity, producing compact scene tokens that condition a pretrained VLA without explicit 3D sensing.
Experiments on RoboTwin~2.0, LIBERO, and a real-robot platform show consistent improvements over prior VLAs and direct multi-view baselines under camera-pose perturbations and varying camera counts.

\textbf{Limitations and broader impacts.}
VersaCamVLA assumes calibrated camera intrinsics and extrinsics and focuses on tabletop bimanual setups with manual camera placement, lacking autonomous viewpoint selection.
Extending the scene-token interface to uncalibrated cameras, active perception that optimizes camera configurations online, and mobile workspaces remains key future work.
In terms of impact, we hope camera-configurable VLA policies lower the deployment barrier across robot platforms by reducing reliance on a fixed camera setup.
However, safety-critical deployment requires further validation against task-specific failures under large camera-configuration shifts.

{\small
\bibliographystyle{unsrtnat}
\bibliography{references}
}

\newpage
\appendix


\section{Benchmark Details}
\label{app:benchmark_details}

This section describes the embodiments, camera pools, demonstrations, and evaluation protocols used in our experiments.
The camera pools provide the available views for scene-token learning; evaluation camera counts follow the configurations reported in the main paper.

\subsection{RoboTwin 2.0}
\label{app:bench_robotwin}
We evaluate on RoboTwin~2.0~\cite{chen2025robotwin} with the Aloha-AgileX bimanual embodiment, selecting 16 fine-grained tasks from the full 50-task suite based on long-horizon and spatial complexity.
The simulator provides 6 environment cameras and 2 wrist-mounted cameras per frame, forming the camera pool from which the scene encoder samples view subsets via Wrist-Augmented Pose Sampling.

The selected tasks are Blocks Ranking RGB, Blocks Ranking Size, Hanging Mug, Move Stapler Pad, Open Laptop, Open Microwave, Pick Diverse Bottles, Place A2B Left, Place A2B Right, Place Burger Fries, Place Dual Shoes, Stack Blocks Three, Stack Bowls Three, Scan Object, Place Empty Cup, and Place Object Basket.
For each task, we collect 50 demonstrations under the \textit{Clean} setting via teleoperation in the official RoboTwin simulator.
VLA models are trained jointly across all 16 tasks, while per-task visuomotor policies are trained separately for each task.

We evaluate under the official \textit{Clean} and domain-randomized (DR) settings, with DR introducing appearance and scene variation.
Each method is evaluated for 100 trials per task per setting, and we report success rates (\%) averaged across the 16 tasks.

\subsection{LIBERO}
\label{app:bench_libero}
We follow the standard LIBERO~\cite{liu2023libero} protocol on a Franka Emika Panda arm in MuJoCo, with the four official task suites.
Our observation setup provides 6 environment cameras and 1 wrist-mounted camera per frame, forming the camera pool for Wrist-Augmented Pose Sampling.
The four suites probe complementary capabilities, with LIBERO-Spatial evaluating spatial relationships among objects, LIBERO-Object testing generalization across object instances, LIBERO-Goal probing language-conditioned goal understanding, and LIBERO-Long covering long-horizon multi-step manipulation.

For each task, we use the 50 human-teleoperated demonstrations provided by the benchmark without modification.
VLA models are trained jointly across all tasks within each suite, while per-task visuomotor policies are trained separately for each task.
Each method is evaluated for 50 trials per task, totaling 500 trials per suite, and we report the average success rate (\%).

\subsection{Real-Robot Platform}
\label{app:bench_real}
We deploy VersaCamVLA on a Cobot Magic (Mobile Aloha) dual-arm platform to evaluate three bimanual manipulation tasks.
The training-time camera setup consists of 4 environment cameras and 2 wrist cameras, providing a flexible pool from which the scene encoder samples view subsets via Wrist-Augmented Pose Sampling.
The three tasks cover sequential bimanual coordination, long-horizon multi-object manipulation, and fine handover-and-insertion.

\textbf{PickCube.} The right arm transfers a red cube to the workspace center, and the left arm picks it up and places it in a bowl.

\textbf{StackCube.} The robot stacks three cubes into a red--green--yellow tower from randomized initial positions, where small placement errors collapse the tower.

\textbf{Insert Test Tube.} The right arm grasps a test tube, hands it to the left arm, and the left arm inserts it into a tube rack.

For each task, we collect 50 human-teleoperated demonstrations.
At evaluation, we run 20 trajectories per task per condition and report the average success rate (\%).

\section{Additional Experimental Results}
\label{app:per_task}

We first examine camera-pose robustness, including unseen viewpoints and calibration noise, and then report camera-count and real-robot results.
Per-task ablations and inference-latency measurements provide further evidence for the design choices discussed in the main paper.

\subsection{Camera-Pose Robustness on RoboTwin~2.0}
\label{app:per_task_robotwin_pose}

\begin{figure}[t]
  \centering
  \includegraphics[width=1.0\linewidth]{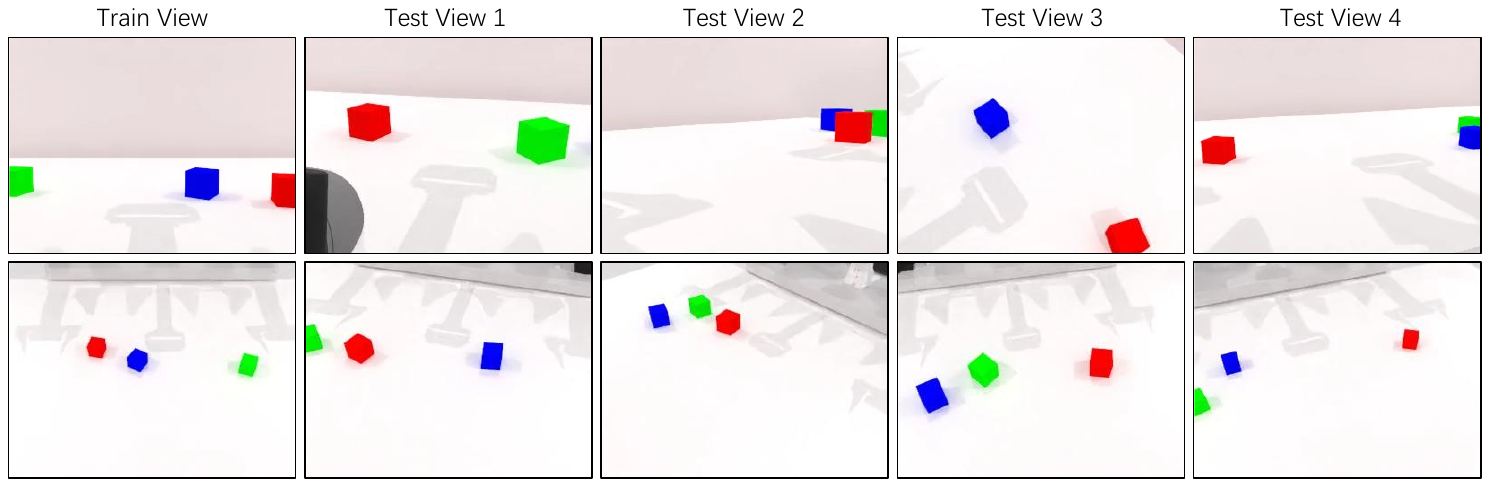}
\caption{
\textbf{Camera-pose perturbations on the RoboTwin~2.0 Stack Blocks Three task.}
The two rows show two of the training camera views.
The first column shows the original, unperturbed camera poses, while the remaining four columns show randomly perturbed camera poses used for robustness evaluation.
}
  \vspace{-5mm}
  \label{fig:robotwin_pose_perturbation}
\end{figure}

Table~\ref{tab:robotwin_pose_per_task} provides the per-task results underlying the RoboTwin~2.0 averages in Table~\ref{tab:camera_pose_ablation}.
Both $\pi_{0.5}$ and VersaCamVLA are evaluated with matched four-view inputs at the trained camera poses (Seen) and at perturbed poses unseen during training (Unseen).
Figure~\ref{fig:robotwin_pose_perturbation} illustrates the viewpoint changes, sampled according to Appendix~\ref{app:impl_perturb}.
The average success rate decreases by $16.00$ percentage points for $\pi_{0.5}$ and $0.81$ percentage points for VersaCamVLA.

\begin{table}[t]
  \centering \small \setlength{\tabcolsep}{4pt}
  \caption{\textbf{Per-task camera-pose robustness on RoboTwin~2.0.} Success rate (\%) at seen and unseen poses under matched four-view inputs. $\Delta$ denotes Unseen minus Seen in percentage points. \textbf{Bold} indicates the higher value between methods for each task and metric, including ties.}
  \label{tab:robotwin_pose_per_task}
  \begin{tabular}{@{}lcccccc@{}}
    \toprule
    \multirow{2}{*}{Task}
    & \multicolumn{3}{c}{$\pi_{0.5}$}
    & \multicolumn{3}{c}{VersaCamVLA} \\
    \cmidrule(lr){2-4} \cmidrule(lr){5-7}
    & Seen & Unseen & $\Delta$ & Seen & Unseen & $\Delta$ \\
    \midrule
    Blocks Ranking RGB    & \textbf{61} & 18 & $-43$ & \textbf{61} & \textbf{56} & \textbf{$-5$} \\
    Blocks Ranking Size   & \textbf{44} & 6 & $-38$ & 31 & \textbf{39} & \textbf{$+8$} \\
    Hanging Mug           & \textbf{14} & \textbf{11} & $-3$ & 6 & 7 & \textbf{$+1$} \\
    Move Stapler Pad      & 8 & 4 & $-4$ & \textbf{9} & \textbf{9} & \textbf{$0$} \\
    Open Laptop           & 90 & 82 & $-8$ & \textbf{96} & \textbf{92} & \textbf{$-4$} \\
    Open Microwave        & 40 & 47 & \textbf{$+7$} & \textbf{60} & \textbf{54} & $-6$ \\
    Pick Diverse Bottles  & 35 & 12 & $-23$ & \textbf{50} & \textbf{50} & \textbf{$0$} \\
    Place A2B Left        & \textbf{49} & 40 & $-9$ & 48 & \textbf{47} & \textbf{$-1$} \\
    Place A2B Right       & 37 & 29 & $-8$ & \textbf{42} & \textbf{45} & \textbf{$+3$} \\
    Place Burger Fries    & 80 & 72 & $-8$ & \textbf{85} & \textbf{85} & \textbf{$0$} \\
    Place Dual Shoes      & 42 & 23 & $-19$ & \textbf{57} & \textbf{63} & \textbf{$+6$} \\
    Place Empty Cup       & 83 & 53 & $-30$ & \textbf{85} & \textbf{72} & \textbf{$-13$} \\
    Place Object Basket   & 64 & 50 & $-14$ & \textbf{65} & \textbf{70} & \textbf{$+5$} \\
    Scan Object           & 20 & 21 & $+1$ & \textbf{39} & \textbf{43} & \textbf{$+4$} \\
    Stack Blocks Three    & 43 & 5  & $-38$ & \textbf{51} & \textbf{43} & \textbf{$-8$} \\
    Stack Bowls Three     & \textbf{72} & 53 & $-19$ & 70 & \textbf{67} & \textbf{$-3$} \\
    \midrule
    Average               & 48.88 & 32.88 & $-16.00$ & \textbf{53.44} & \textbf{52.63} & \textbf{$-0.81$} \\
    \bottomrule
  \end{tabular}
  \vspace{-5mm}
\end{table}

\subsection{Camera-Pose Robustness on LIBERO}
\label{app:per_task_libero_pose}
Table~\ref{tab:libero_pose_per_suite} extends the camera-pose evaluation to the four LIBERO suites with four-view inputs.
The same VersaCamVLA model is evaluated at seen and unseen camera poses; $\pi_0$ and $\pi_{0.5}$ are reported at seen poses for reference.
VersaCamVLA's average success decreases from $95.70\%$ to $95.15\%$, a drop of $0.55$ percentage points, consistent with the robustness observed on RoboTwin~2.0.

\begin{table}[t]
  \centering \small \setlength{\tabcolsep}{6pt}
  \caption{\textbf{Camera-pose robustness on LIBERO.} Success rate (\%) within each official suite under four-view inputs. Seen denotes trained camera poses and Unseen denotes perturbed poses. \textbf{Bold} indicates the highest value in each column, including ties.}
  \label{tab:libero_pose_per_suite}
  \begin{tabular}{@{}lccccc@{}}
    \toprule
    Method (Setup) & Spatial & Goal & Object & Long & Avg \\
    \midrule
    $\pi_0$ (Seen)         & 94.8 & 87.0 & 97.0 & 83.6 & 90.6 \\
    $\pi_{0.5}$ (Seen)     & 96.8 & 88.2 & 96.6 & 82.4 & 91.0 \\
    VersaCamVLA (Seen)     & \textbf{98.2} & 95.0 & \textbf{97.8} & \textbf{91.8} & \textbf{95.7} \\
    VersaCamVLA (Unseen)   & \textbf{98.2} & \textbf{96.8} & 95.8 & 89.8 & 95.15 \\
    \bottomrule
  \end{tabular}
  \vspace{-5mm}
\end{table}

\subsection{Generalization to Challenging Unseen Camera Pose Shifts}
\label{app:large_pose_shifts}

The unseen-pose results in Table~\ref{tab:camera_pose_ablation} average over a broad range of camera configurations, leaving the effect of viewpoint-shift magnitude unresolved.
To examine robustness as the shift increases, we stratify the RoboTwin~2.0 evaluation by absolute azimuth offset $|\Delta\phi|$.
The original protocol samples $\Delta\phi\sim\mathcal{U}[-90^\circ,90^\circ]$, as detailed in Appendix~\ref{app:impl_perturb}.
Here, we restrict sampling to three absolute-offset intervals, $[0^\circ,30^\circ]$, $[30^\circ,60^\circ]$, and $[60^\circ,90^\circ]$, with the last interval probing the largest viewpoint shifts.
All other evaluation settings follow Table~\ref{tab:camera_pose_ablation}, including matched four-view inputs and the radial and pitch perturbations described in Appendix~\ref{app:impl_perturb}.

\begin{table}[htbp]
  \centering
  \small
  \setlength{\tabcolsep}{4pt}
  \caption{\textbf{Generalization across unseen camera-pose shift magnitudes on RoboTwin~2.0.} Average success rate (\%) at the seen pose and under perturbations grouped by absolute azimuth offset. \textbf{Bold} indicates the best result in each column.}
  \label{tab:azimuth_shift_robustness}
  \begin{tabular*}{\linewidth}{@{\extracolsep{\fill}}lcccc@{}}
    \toprule
    \multirow{2}{*}{Method} & \multirow{2}{*}{Seen Pose} & \multicolumn{3}{c}{Unseen Pose: $|\Delta\phi|$} \\
    \cmidrule(lr){3-5}
    & & $[0^\circ,30^\circ]$ & $[30^\circ,60^\circ]$ & $[60^\circ,90^\circ]$ \\
    \midrule
    $\pi_{0.5}$ & 48.88 & 43.38 & 34.13 & 30.06 \\
    VersaCamVLA & \textbf{53.44} & \textbf{53.25} & \textbf{52.13} & \textbf{53.38} \\
    \bottomrule
  \end{tabular*}
\end{table}

As shown in Table~\ref{tab:azimuth_shift_robustness}, VersaCamVLA maintains stable performance across all three intervals, with a maximum drop of $1.31$ percentage points relative to the seen pose.
In contrast, $\pi_{0.5}$ becomes progressively less reliable as the viewpoint shift increases, falling from $48.88\%$ to $30.06\%$ under the largest azimuth offsets, a drop of $18.82$ percentage points.
The performance gap therefore widens from $4.56$ percentage points at the seen pose to $23.32$ percentage points in the $[60^\circ,90^\circ]$ interval.
These findings show that VersaCamVLA's robustness extends to large unseen viewpoint shifts within the evaluated range.

\subsection{Robustness under Camera Calibration Noise}
\label{app:calibration_noise}

VersaCamVLA uses camera extrinsics to encode the spatial relationships among input views, making calibration accuracy a practical consideration for deployment.
To evaluate sensitivity to calibration errors, we perturb the camera extrinsics supplied to the model during evaluation on RoboTwin~2.0.
At the beginning of each episode, we independently sample Gaussian translation and rotation perturbations for each camera and keep them fixed throughout the episode, simulating a static extrinsic-calibration bias.
The parameters $\sigma_t$ and $\sigma_r$ denote the per-axis standard deviations of the translation and rotation perturbations, respectively.

\begin{table}[htbp]
  \centering
  \small
  \setlength{\tabcolsep}{6pt}
  \caption{\textbf{Robustness to camera extrinsic-calibration noise on RoboTwin~2.0.} Average success rate (\%) under Gaussian perturbations fixed within each episode. Translation and rotation noise levels are specified by their per-axis standard deviations.}
  \label{tab:calibration_noise}
  \begin{tabular*}{\linewidth}{@{\extracolsep{\fill}}lccc@{}}
    \toprule
    Extrinsic-noise level & $\sigma_t$ (mm) & $\sigma_r$ (degrees) & Average SR (\%) \\
    \midrule
    Clean    & 0  & 0 & 52.56 \\
    Moderate & 20 & 2 & 52.13 \\
    Strong   & 50 & 5 & 50.38 \\
    \bottomrule
  \end{tabular*}
\end{table}

Table~\ref{tab:calibration_noise} shows that moderate calibration noise ($\sigma_t=20$~mm, $\sigma_r=2^\circ$) reduces average success from $52.56\%$ to $52.13\%$, a drop of only $0.43$ percentage points.
Under stronger noise ($\sigma_t=50$~mm, $\sigma_r=5^\circ$), VersaCamVLA retains a success rate of $50.38\%$, with a drop of $2.18$ percentage points from the clean setting.
These results indicate that the scene-token interface tolerates moderate inaccuracies in camera extrinsics and degrades gradually over the evaluated noise levels.

\subsection{Camera-Count Flexibility on RoboTwin~2.0}
\label{app:per_task_robotwin_count}

Table~\ref{tab:robotwin_count_per_task} expands the RoboTwin~2.0 results in Table~\ref{tab:camera_count_ablation}, using a single VersaCamVLA model at 3, 4, 5, and 6 input views.
The corresponding $\pi_{0.5}$ baselines are trained separately at three and four views, reaching $35.88\%$ and $48.88\%$, respectively; five- and six-view baselines are not evaluated.
VersaCamVLA maintains average success between $52.56\%$ and $55.63\%$ across the evaluated camera counts, with the highest average at six views.

\begin{table}[t]
  \centering \small \setlength{\tabcolsep}{4pt}
  \caption{\textbf{Per-task camera-count variation on RoboTwin~2.0 \textit{Clean}.} Success rate (\%) for a single VersaCamVLA model evaluated with 3, 4, 5, and 6 views (V). \textbf{Bold} indicates the highest value in each row, including ties.}
  \label{tab:robotwin_count_per_task}
  \begin{tabular}{@{}lcccc@{}}
    \toprule
    Task & 3V & 4V & 5V & 6V \\
    \midrule
    Blocks Ranking RGB    & \textbf{67} & 61 & 61 & \textbf{67} \\
    Blocks Ranking Size   & 35 & 31 & \textbf{39} & 35 \\
    Hanging Mug           & \textbf{10} & 6 & 9 & 3 \\
    Move Stapler Pad      & 8 & 9 & \textbf{13} & 4 \\
    Open Laptop           & \textbf{96} & \textbf{96} & 94 & \textbf{96} \\
    Open Microwave        & 51 & 60 & 57 & \textbf{67} \\
    Pick Diverse Bottles  & 47 & \textbf{50} & 49 & 41 \\
    Place A2B Left        & 54 & 48 & 49 & \textbf{56} \\
    Place A2B Right       & 45 & 42 & 44 & \textbf{48} \\
    Place Burger Fries    & 81 & 85 & 83 & \textbf{88} \\
    Place Dual Shoes      & 51 & 57 & 54 & \textbf{58} \\
    Place Empty Cup       & 81 & \textbf{85} & 78 & \textbf{85} \\
    Place Object Basket   & 76 & 65 & 70 & \textbf{79} \\
    Scan Object           & 35 & 39 & 40 & \textbf{45} \\
    Stack Blocks Three    & 43 & \textbf{51} & 45 & 48 \\
    Stack Bowls Three     & 61 & \textbf{70} & 65 & \textbf{70} \\
    \midrule
    Average               & 52.56 & 53.44 & 53.13 & \textbf{55.63} \\
    \bottomrule
  \end{tabular}
\end{table}

\subsection{Real-Robot Camera-Pose and Camera-Count Results}
\label{app:per_task_real_robot}
Tables~\ref{tab:real_pose_per_task} and~\ref{tab:real_count_per_task} provide the per-task real-robot results underlying Tables~\ref{tab:camera_pose_ablation} and~\ref{tab:camera_count_ablation}, respectively.
The same VersaCamVLA model is evaluated at seen and unseen poses with four views, and with an additional fifth view at test time.
The $\pi_{0.5}$ baselines are trained separately for the three- and four-view configurations.
Evaluation follows Appendix~\ref{app:bench_real}, with 20 trajectories per task and condition.

\begin{table}[t]
  \centering \small \setlength{\tabcolsep}{6pt}
  \caption{\textbf{Per-task camera-pose robustness on the real robot.} Success rate (\%) under matched four-view inputs. $\Delta$ denotes Unseen minus Seen in percentage points. \textbf{Bold} indicates the higher value between methods for each task and metric, including ties.}
  \label{tab:real_pose_per_task}
  \begin{tabular}{@{}lcccccc@{}}
    \toprule
    \multirow{2}{*}{Task}
    & \multicolumn{3}{c}{$\pi_{0.5}$}
    & \multicolumn{3}{c}{VersaCamVLA} \\
    \cmidrule(lr){2-4} \cmidrule(lr){5-7}
    & Seen & Unseen & $\Delta$ & Seen & Unseen & $\Delta$ \\
    \midrule
    PickCube         & 70 & 35 & $-35$ & \textbf{85}    & \textbf{80}             & \textbf{$-5$} \\
    StackCube        & \textbf{55} & 5  & $-50$ & \textbf{55}             & \textbf{60}    & \textbf{$+5$} \\
    Insert Test Tube & 0  & 0  & $0$   & \textbf{20}             & \textbf{35}    & \textbf{$+15$} \\
    \midrule
    Average          & 41.67 & 13.33 & $-28.33$ & \textbf{53.33} & \textbf{58.33} & \textbf{$+5.00$} \\
    \bottomrule
  \end{tabular}
\end{table}
\begin{table}[t]
  \centering \small \setlength{\tabcolsep}{6pt}
  \caption{\textbf{Per-task camera-count variation on the real robot.} Success rate (\%) for $\pi_{0.5}$ trained separately at 3 and 4 views and a single VersaCamVLA model evaluated at 4 and 5 views (V). \textbf{Bold} indicates the highest value in each row, including ties.}
  \label{tab:real_count_per_task}
  \begin{tabular}{@{}lcccc@{}}
    \toprule
    \multirow{2}{*}{Task}
    & \multirow{2}{*}{$\pi_{0.5}$ (3V)}
    & \multirow{2}{*}{$\pi_{0.5}$ (4V)}
    & \multicolumn{2}{c}{VersaCamVLA} \\
    \cmidrule(lr){4-5}
    & & & 4V & 5V \\
    \midrule
    PickCube         & 50    & 70    & \textbf{85}    & \textbf{85} \\
    StackCube        & 40    & 55    & 55             & \textbf{65} \\
    Insert Test Tube & 5     & 0     & 20             & \textbf{45} \\
    \midrule
    Average          & 31.67 & 41.67 & 53.33          & \textbf{65.00} \\
    \bottomrule
  \end{tabular}
\end{table}

\subsection{Per-Task Ablation Results}
\label{app:per_task_ablation}
Tables~\ref{tab:ablation_token_per_task} and~\ref{tab:ablation_encoder_per_task} expand the token-count and spatial-encoder ablations in Tables~\ref{tab:token_ablation_robotwin} and~\ref{tab:encoder_ablation_robotwin}, respectively, on RoboTwin~2.0 \textit{Clean}.
Table~\ref{tab:ablation_signal_per_task} provides the per-task comparison between RGB-only and full multi-signal supervision from Table~\ref{tab:ablation_signal_summary} under RoboTwin~2.0 \textit{DR}.
The full supervision setting combines RGB, semantic, and edge targets.
\begin{table}[t]
  \centering \small \setlength{\tabcolsep}{6pt}
  \caption{\textbf{Per-task ablation of compressed scene-token count.} Success rate (\%) on RoboTwin~2.0 \textit{Clean}. Token counts refer to the compressed scene tokens passed to the policy. \textbf{Bold} indicates the highest value in each row, including ties.}
  \label{tab:ablation_token_per_task}
  \begin{tabular}{@{}lccc@{}}
    \toprule
    Task & 48 tokens & 192 tokens & 432 tokens \\
    \midrule
    Blocks Ranking RGB    & 55 & \textbf{67} & 55 \\
    Blocks Ranking Size   & \textbf{40} & 35 & 32 \\
    Hanging Mug           & 8 & 3 & \textbf{16} \\
    Move Stapler Pad      & \textbf{17} & 4 & 13 \\
    Open Laptop           & 89 & \textbf{96} & 92 \\
    Open Microwave        & 20 & \textbf{67} & 61 \\
    Pick Diverse Bottles  & \textbf{44} & 41 & \textbf{44} \\
    Place A2B Left        & \textbf{56} & \textbf{56} & 42 \\
    Place A2B Right       & \textbf{52} & 48 & 43 \\
    Place Burger Fries    & 84 & \textbf{88} & 52 \\
    Place Dual Shoes      & 44 & \textbf{58} & 40 \\
    Place Empty Cup       & \textbf{87} & 85 & 84 \\
    Place Object Basket   & 75 & \textbf{79} & 67 \\
    Scan Object           & 36 & \textbf{45} & 29 \\
    Stack Blocks Three    & 34 & \textbf{48} & 29 \\
    Stack Bowls Three     & 60 & \textbf{70} & 61 \\
    \midrule
    Average               & 50.06 & \textbf{55.63} & 47.50 \\
    \bottomrule
  \end{tabular}
\end{table}
\begin{table}[t]
  \centering \small \setlength{\tabcolsep}{6pt}
  \caption{\textbf{Per-task ablation of spatial encoder architecture.} Success rate (\%) on RoboTwin~2.0 \textit{Clean} with pooling or convolutional compression. \textbf{Bold} indicates the highest value in each row, including ties.}
  \label{tab:ablation_encoder_per_task}
  \begin{tabular}{@{}lcc@{}}
    \toprule
    Task & Pooling & Conv \\
    \midrule
    Blocks Ranking RGB    & 61 & \textbf{67} \\
    Blocks Ranking Size   & 26 & \textbf{35} \\
    Hanging Mug           & \textbf{11} & 3 \\
    Move Stapler Pad      & \textbf{9} & 4 \\
    Open Laptop           & 91 & \textbf{96} \\
    Open Microwave        & 52 & \textbf{67} \\
    Pick Diverse Bottles  & \textbf{48} & 41 \\
    Place A2B Left        & 51 & \textbf{56} \\
    Place A2B Right       & \textbf{53} & 48 \\
    Place Burger Fries    & \textbf{88} & \textbf{88} \\
    Place Dual Shoes      & 49 & \textbf{58} \\
    Place Empty Cup       & \textbf{87} & 85 \\
    Place Object Basket   & 76 & \textbf{79} \\
    Scan Object           & 27 & \textbf{45} \\
    Stack Blocks Three    & 17 & \textbf{48} \\
    Stack Bowls Three     & 56 & \textbf{70} \\
    \midrule
    Average               & 50.13 & \textbf{55.63} \\
    \bottomrule
  \end{tabular}
  \vspace{-5mm}
\end{table}
\begin{table}[t]
  \centering \small \setlength{\tabcolsep}{6pt}
  \caption{\textbf{Per-task ablation of scene-token supervision.} Success rate (\%) on RoboTwin~2.0 \textit{DR} with RGB-only or full multi-signal supervision. \textbf{Bold} indicates the highest value in each row, including ties.}
  \label{tab:ablation_signal_per_task}
  \begin{tabular}{@{}lcc@{}}
    \toprule
    Task & RGB only & Multi-signal \\
    \midrule
    Blocks Ranking RGB    & 13 & \textbf{14} \\
    Blocks Ranking Size   & 6 & \textbf{9} \\
    Hanging Mug           & 6 & \textbf{11} \\
    Move Stapler Pad      & 3 & \textbf{7} \\
    Open Laptop           & \textbf{67} & 63 \\
    Open Microwave        & 12 & \textbf{54} \\
    Pick Diverse Bottles  & \textbf{13} & \textbf{13} \\
    Place A2B Left        & 15 & \textbf{20} \\
    Place A2B Right       & \textbf{22} & 19 \\
    Place Burger Fries    & 55 & \textbf{77} \\
    Place Dual Shoes      & 24 & \textbf{25} \\
    Place Empty Cup       & 2 & \textbf{41} \\
    Place Object Basket   & \textbf{33} & 30 \\
    Scan Object           & 13 & \textbf{18} \\
    Stack Blocks Three    & 26 & \textbf{35} \\
    Stack Bowls Three     & 25 & \textbf{36} \\
    \midrule
    Average               & 20.94 & \textbf{29.50} \\
    \bottomrule
  \end{tabular}
  \vspace{-5mm}
\end{table}

\subsection{Latency and Efficiency}
\label{sec:exp_latency}
\begin{figure}[h]
  \centering
  \includegraphics[width=0.55\linewidth]{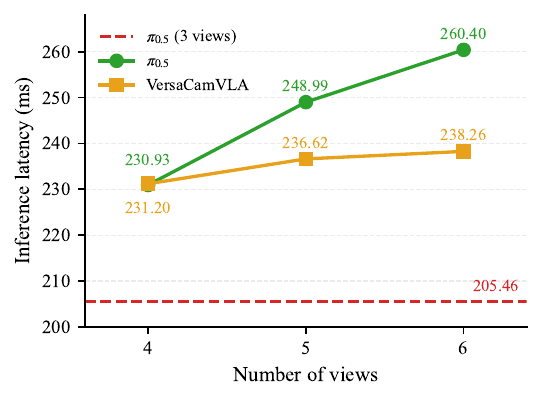}
  \caption{\textbf{Inference latency with varying camera counts.} Measurements use a single NVIDIA 4090 at batch size 1. The red dashed line marks the three-view $\pi_{0.5}$ reference. VersaCamVLA incurs a smaller latency increase as additional input views are incorporated.}
  \vspace{-5mm}
  \label{fig:latency}
\end{figure}

Figure~\ref{fig:latency} reports inference latency on a single NVIDIA 4090 at batch size 1 as the number of input views varies.
At four views, VersaCamVLA takes $231.20$~ms per inference step, compared with $230.93$~ms for $\pi_{0.5}$, a difference of $0.27$~ms.
Increasing the input count from four to six views adds $29.47$~ms for $\pi_{0.5}$ and $7.06$~ms for VersaCamVLA, yielding a latency gap of $22.14$~ms at six views.
This smaller increase is consistent with the fixed-size scene-token interface: additional views are processed by the scene encoder while the number of supplementary tokens passed to the policy remains fixed.

\section{Implementation Details}
\label{app:impl_details}

This section specifies the architecture, the two training stages, and the camera-pose sampling protocol used in the experiments.

\subsection{Architecture Hyperparameters}
\label{app:impl_arch}

\paragraph{Tokenization.}
RGB observations are resized to $224\times224$ and tokenized with patch size $p=8$, yielding $28\times28=784$ patches per view.
Source-view inputs are 9-channel (3 RGB + 3 Pl\"ucker direction + 3 Pl\"ucker moment), while target-view query inputs are 6-channel (3 Pl\"ucker direction + 3 moment), since target views provide only ray geometry.

\paragraph{Scene encoder and decoder.}
Both the scene encoder $\mathrm{Enc}_\theta$ and the scene decoder $\mathrm{Dec}_\psi$ are 12-layer Transformers with hidden dimension $d=768$, head dimension $64$ (12 attention heads), QK-normalized attention, and depth-aware initialization for stable training.
The scene encoder produces $L=3072$ learnable scene tokens organized as a $3\times32\times32$ spatial layout, providing a fixed-size summary of the source views regardless of the input camera count.

\paragraph{Spatial encoder $g_\eta$.}
The spatial encoder is a lightweight convolutional module that maps the learned scene tokens to the token format expected by the base VLA policy.
Specifically, it compresses the original $3\times32\times32$ scene-token grid with channel dimension $768$ into a compact $3\times8\times8$ representation with channel dimension $2048$, reducing the token count from $3072$ to $192$.
This design provides a task-oriented bottleneck that preserves useful spatial cues while avoiding excessive token length during policy learning.
In practice, $g_\eta$ is implemented with standard convolutional operations, including channel projection, local aggregation, spatial downsampling, and output projection.

\paragraph{Multi-signal prediction heads.}
The RGB, semantic, and edge heads $h_{\rm rgb}, h_{\rm sem}, h_{\rm edge}$ are lightweight convolutional networks applied to the target-view decoder output $\mathbf{Y}^{c}$, each producing a full-resolution prediction at the target view.
All three heads are discarded together with the scene decoder at the end of Stage~I and do not appear at deployment.

\subsection{Stage I: Scene-Token Interface Learning}
\label{app:impl_stage1}

\paragraph{Loss formulation.}
Stage~I minimizes the multi-signal target-view loss $\mathcal{L}_{\rm scene} = \mathcal{L}_{\rm rgb} + \lambda_{\rm sem}\,\mathcal{L}_{\rm sem} + \lambda_{\rm edge}\,\mathcal{L}_{\rm edge}$, where $\mathcal{L}_{\rm rgb}$ combines pixel MSE with a perceptual term weighted by $\lambda_{\rm perc}$, and $\mathcal{L}_{\rm sem}$ and $\mathcal{L}_{\rm edge}$ are MSE losses on the semantic and edge predictions.
We use $\lambda_{\rm sem} = \lambda_{\rm edge} = 1.0$ and $\lambda_{\rm perc} = 0.5$, balancing structure-aware supervision against photometric reconstruction.
The supervision targets are described in Appendix~\ref{app:qual_supervision}.

\paragraph{Optimization.}
Stage~I is optimized with AdamW ($\beta_1{=}0.9$, $\beta_2{=}0.95$, weight decay $0.05$) at a constant learning rate of $1\times10^{-4}$ after a $3{,}000$-step linear warmup, with gradient clipping at norm $1.0$.
Training runs for $100{,}000$ steps in bf16 mixed precision (with TF32 matmul) on $8\times$NVIDIA H100 GPUs at a total batch size of $208$.
On RoboTwin~2.0, each scene provides eight available views.
WAPS samples the source subset independently at each frame, while the decoder predicts RGB, semantic, and edge signals at all available target views, following Sec.~\ref{sec:camera-sampling}.
The resulting objective includes both reconstruction of source views and prediction of views omitted from the source subset.

\paragraph{Initialization.}
The scene encoder and decoder are initialized from a general LVSM-style scene encoder--decoder checkpoint and fine-tuned on RoboTwin~2.0 to specialize to manipulation scenes, while the multi-signal prediction heads ($h_{\rm rgb}, h_{\rm sem}, h_{\rm edge}$) are randomly initialized.

\subsection{Multi-Signal Supervision Targets}
\label{app:qual_supervision}
\begin{figure}[h]
  \centering
  \includegraphics[width=\linewidth]{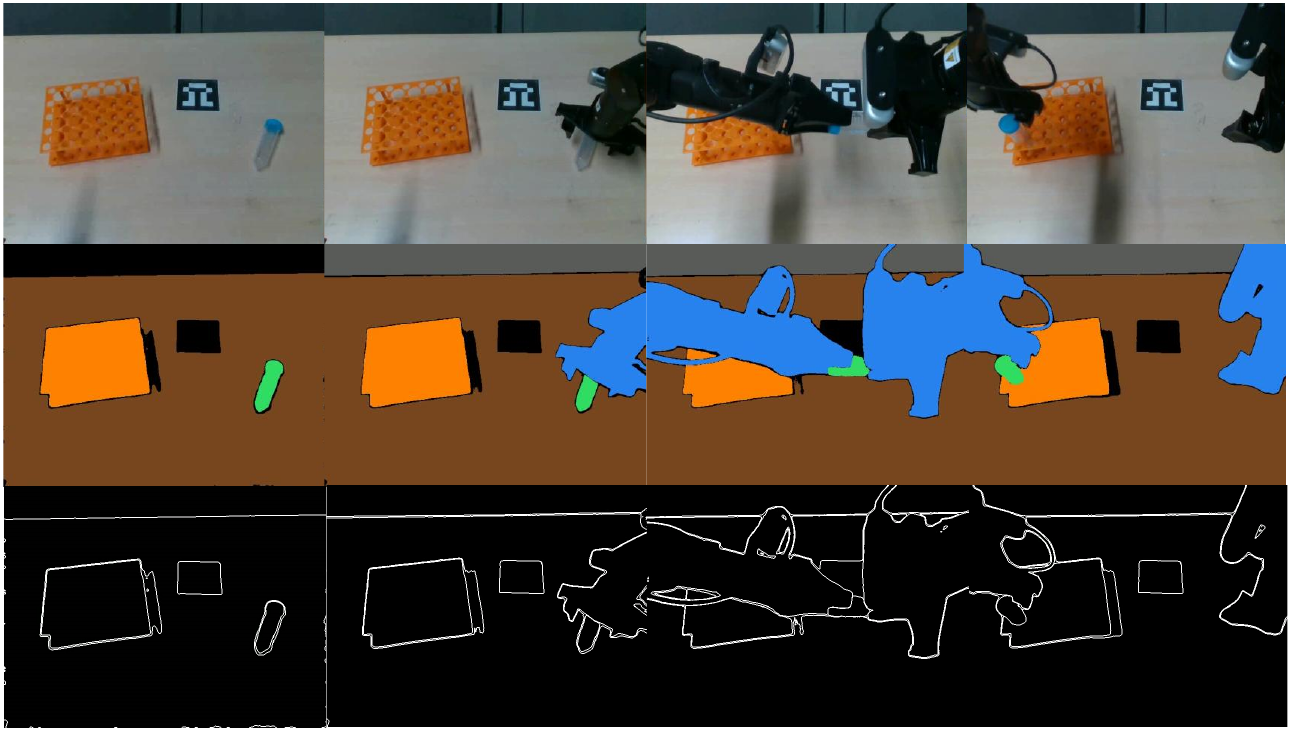}
  \caption{
    \textbf{Multi-signal supervision targets on the real-robot \textit{Insert Test Tube} task.}
    Rows (top to bottom) show the raw RGB observation, the semantic mask produced by a SAM3 segmenter fine-tuned on our real-robot data, and the Canny edge map.
    Columns (left to right) span four key task states, from the right arm grasping the test tube to the left arm completing insertion into the rack.
  }
  \label{fig:sam3_targets}
  \vspace{-5mm}
\end{figure}

Stage~I uses RGB observations together with semantic and edge targets.
Semantic targets come from the simulator's segmentation sensor in simulation and from a SAM3~\cite{carion2025sam3} segmenter fine-tuned on our real-robot data for real observations; edge targets are extracted with a Canny detector.
Figure~\ref{fig:sam3_targets} illustrates all three channels on the real-robot \textit{Insert Test Tube} task.

\subsection{Stage II: Scene-Token-Conditioned Policy Learning}
\label{app:impl_stage2}

\paragraph{Loss formulation.}
Stage~II minimizes the flow-matching action regression loss $\mathcal{L}_{\rm action}$, predicting an action chunk of horizon $H_a = 50$ from the language instruction, native RGB observations, supplementary scene tokens $\bar{\mathbf{Z}}_t$, and proprioceptive state.
The scene encoder is frozen throughout this stage, and the spatial encoder $g_\eta$ is jointly optimized with the base VLA's trainable parameters under $\mathcal{L}_{\rm action}$, so that $g_\eta$ acts as a task-driven information bottleneck retaining scene-token information useful for action prediction.

\paragraph{Optimization.}
Stage~II is optimized with AdamW at a learning rate of $5\times10^{-5}$ for $30{,}000$ steps in bf16 mixed precision on $8\times$NVIDIA H100 GPUs at a total batch size of $512$.

\paragraph{Initialization.}
The base VLA is initialized from the pretrained $\pi_{0.5}$-base weights~\cite{intelligence2025pi_05}, while the spatial encoder $g_\eta$ is randomly initialized.
The scene encoder $\mathrm{Enc}_\theta$ is loaded from the Stage~I checkpoint (Appendix~\ref{app:impl_stage1}) and remains frozen throughout Stage~II.

\subsection{Camera-Pose Perturbation Sampling}
\label{app:impl_perturb}
At evaluation time, we perturb each camera by sampling its pose on a spherical shell centered on the workspace.
For each camera, we independently draw an azimuth offset $\Delta\phi\sim\mathcal{U}[-90^\circ, 90^\circ]$, a radial scale $r\sim\mathcal{U}[0.8, 1.2]$ that rescales the camera's distance to the workspace, and a pitch $\theta\sim\mathcal{U}[25^\circ, 75^\circ]$; the camera is then reoriented to look at the workspace center.
Camera configurations are resampled for each trial and shared between $\pi_{0.5}$ and VersaCamVLA to ensure a matched comparison.
The stratified evaluation in Appendix~\ref{app:large_pose_shifts} restricts the absolute azimuth offset to the specified interval while retaining the remaining sampling settings.

\section{Technical appendices and supplementary material}
Technical appendices with additional results, figures, graphs, and proofs may be submitted with the paper submission before the full submission deadline (see above). You can upload a ZIP file for videos or code, but do not upload a separate PDF file for the appendix. There is no page limit for the technical appendices. 

Note: Think of the appendix as ``optional reading'' for reviewers. The paper must be able to stand alone without the appendix; for example, adding critical experiments that support the main claims to an appendix is inappropriate. 



\end{document}